\documentclass[10pt,twocolumn,letterpaper]{article}

\usepackage{cvpr}              
\usepackage{xcolor}

\usepackage{lipsum}
\usepackage{listings}
\usepackage{subcaption}
\usepackage{ragged2e}
\usepackage{array}
\usepackage{times}
\usepackage{epsfig}
\usepackage{graphicx}
\usepackage{float}
\usepackage{marvosym}
\usepackage{wrapfig}
\usepackage{amsmath,amssymb,amsthm}
\usepackage{algorithm,algorithmicx,algpseudocode}
\usepackage{tcolorbox}
\usepackage{bm,xspace}
\usepackage{comment}
\usepackage{multirow}
\usepackage{balance}
\usepackage{url}
\usepackage{booktabs}
\usepackage{etoolbox,siunitx}
\usepackage{calc}
\usepackage{pifont,hologo}
\usepackage{color}
\usepackage{adjustbox}
\usepackage{amsmath}
\usepackage{enumitem}
\usepackage[normalem]{ulem}  %
\usepackage{contour}
\usepackage{colortbl} 
\usepackage{soul}%
\usepackage{tocloft} %
\usepackage{etoc} %
\PassOptionsToPackage{dvipsnames}{xcolor}
\usepackage[pagebackref,breaklinks,colorlinks,linkcolor=link,citecolor=cite]{hyperref}
\usepackage{natbib}
\PassOptionsToPackage{square,numbers,comma,sort}{natbib}

\definecolor{mygreen}{HTML}{d0f4de}
\definecolor{myblue}{HTML}{50ACE9}
\newlength\savewidth

\definecolor{Highlight}{HTML}{39b54a}

\definecolor{blue}{HTML}{004bb3}
\definecolor{red}{HTML}{cc1100}
\definecolor{orange}{HTML}{cc7700}
\definecolor{gray}{HTML}{efefef}
\definecolor{darkgreen}{HTML}{228B22}
\definecolor{darkgray}{HTML}{757575}

\definecolor{cite}{HTML}{3270b5}
\definecolor{link}{HTML}{b53532}
\definecolor{link}{HTML}{cc1100}
\definecolor{scratch}{HTML}{001219}
\definecolor{pretrain}{HTML}{0A9396}

\contourlength{0.8pt}

\renewcommand{\eqref}[1]{Eq.~\ref{#1}}

\newcolumntype{x}[1]{>{\centering\arraybackslash}p{#1}}
\newcolumntype{y}[1]{>{\raggedright\arraybackslash}p{#1}}
\newcolumntype{z}[1]{>{\raggedleft\arraybackslash}p{#1}}

\DeclareMathSymbol{@}{\mathord}{letters}{"3B}

\makeatletter
\DeclareRobustCommand\onedot{\futurelet\@let@token\@onedot}
\def\@onedot{\ifx\@let@token.\else.\null\fi\xspace}

\newcommand*{\Rom}[1]{\expandafter\@slowromancap\romannumeral #1@}
\newcommand*{\rom}[1]{\expandafter\romannumeral #1}

\def\1{\bm{1}}

\DeclareMathAlphabet{\mathsfit}{\encodingdefault}{\sfdefault}{m}{sl}
\SetMathAlphabet{\mathsfit}{bold}{\encodingdefault}{\sfdefault}{bx}{n}

\let\originalleft\left
\let\originalright\right
\renewcommand{\left}{\mathopen{}\mathclose\bgroup\originalleft}
\renewcommand{\right}{\aftergroup\egroup\originalright}

\definecolor{cvprblue}{rgb}{0.21,0.49,0.74}
\usepackage{listings}
\usepackage{xcolor}
\lstdefinelanguage{json}{
    basicstyle=\ttfamily\small,
    numbers=left,
    numberstyle=\scriptsize\color{gray},
    stepnumber=1,
    numbersep=8pt,
    showstringspaces=false,
    breaklines=true,
    frame=lines,
    backgroundcolor=\color{white},
    literate=
     *{0}{{{\color{blue}0}}}{1}
      {1}{{{\color{blue}1}}}{1}
      {2}{{{\color{blue}2}}}{1}
      {3}{{{\color{blue}3}}}{1}
      {4}{{{\color{blue}4}}}{1}
      {5}{{{\color{blue}5}}}{1}
      {6}{{{\color{blue}6}}}{1}
      {7}{{{\color{blue}7}}}{1}
      {8}{{{\color{blue}8}}}{1}
      {9}{{{\color{blue}9}}}{1}
      {:}{{{\color{red}:}}}{1}
      {,}{{{\color{red},}}}{1}
      {\{}{{{\color{black}\{}}}{1}
      {\}}{{{\color{black}\}}}}{1}
      {[}{{{\color{black}[}}}{1}
      {]}{{{\color{black}]}}}{1},
}

\def\paperID{4345} 
\def\confName{CVPR}
\def\confYear{2025}

\title{STEVE: A Step Verification Pipeline for Computer-use Agent Training\thanks{Corresponding author: Shu Liu}}

\author{Fanbin Lu$^{1}$ \hspace{1pt} Zhisheng Zhong$^{1}$ \hspace{1pt} Ziqin Wei$^{1}$ \hspace{1pt} Shu Liu$^{2*}$ \hspace{1pt} Chi-Wing Fu$^1$ \hspace{1pt} Jiaya Jia$^{2,3}$
\\
\\
CUHK$^1$ \hspace{2pt} SmartMore$^2$ \hspace{2pt} HKUST$^3$ 
}

\begin{document}
\maketitle
\begin{abstract}
Developing AI agents to autonomously manipulate graphical user interfaces is a long challenging task.
Recent advances in data scaling law inspire us to train computer-use agents with a scaled instruction set, yet 
using behavior cloning to train agents still requires immense high-quality trajectories. 
To meet the scalability need, we design STEVE, a step verification pipeline for computer-use agent training.
First, we establish a large instruction set for computer-use agents and collect trajectory data with some suboptimal agents. 
GPT-4o is used to verify the correctness of each step in the trajectories based on the screens before and after the action execution, assigning each step with a binary label. 
Last, we adopt the Kahneman \& Tversky Optimization to optimize the agent from the binary stepwise labels.
Extensive experiments manifest that our agent outperforms supervised finetuning by leveraging both positive and negative actions within a trajectory.
Also, STEVE enables us to train a 7B vision-language model as a computer-use agent, achieving leading performance in the challenging live desktop environment WinAgentArena with great efficiency at a reduced cost.
Code and data: \url{https://github.com/FanbinLu/STEVE}

\end{abstract}    
\section{Introduction}
\label{sec:intro}
Creating AI agents that act like humans to manipulate graphical user interfaces (GUIs) is a longstanding but very challenging goal in artificial intelligence. 
%
Given the increasing need of performing tasks on digital devices, the potential to enhance productivity by deploying AI agents to automate complex and repetitive operations is immense.
Recent progress in large vision-language models (VLMs), such as GPT-4o, showcases exceptional capabilities in natural language understanding, reasoning, and visual perception~\cite{liu2024visual, wang2024qwen2}. These advances open new possibilities for designing AI agents to interact with GUIs similar to how humans do. 
However, to achieve these capabilities
still involves significant challenges that need to be addressed.

\begin{figure}[th!]
\centering
    \includegraphics[width=0.9\linewidth]{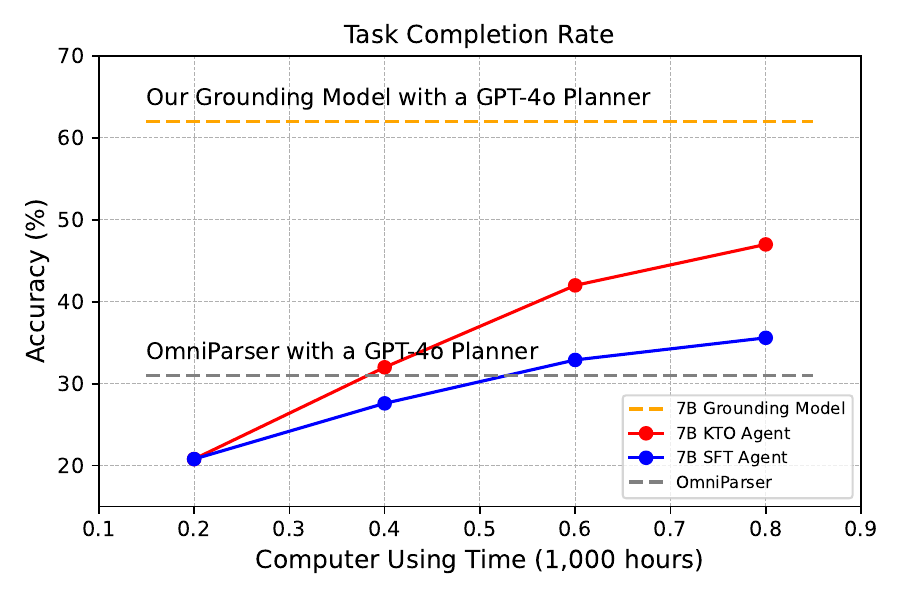}
    \caption{Windows File Explorer task completion rate of different computer-use agents:
    (i) Our powerful GUI grounding model achieves the current best task completion rate, setting a promising upper bound for computer-use agent finetuning. 
    (ii) Using STEVE, our step verification pipeline, we are able to train 
    our agents with KTO (red), which consistently outperforms (iii) the supervised finetuning (SFT).
    Notably, with increased computer operating time (x-axis), our 7B KTO agent is able to outperform the OmniParser with the GPT-4o planner. }
    %
    
\label{fig:teaser}
\end{figure}

One of the primary challenges lies in the precise understanding and localization of UI elements on the screen. The high-resolution displays and complicated modern GUI pattern challenge the agent's ability to correctly interact with the device. Traditional detection and OCR approaches~\cite{xie2020uied, altinbas2022gui} fall short of understanding the functionalities of UI components, necessitating a large VLM to support this task.
Another significant challenge is the planning and execution of multi-step tasks that often involve long sequences of actions, thus highly demanding the agent's long-term and dynamic planning capabilities. Real-world desktop environments~\cite{xie2024osworld, bonatti2024windows} have been proposed to evaluate the multi-step planning and complex-task-solving ability.

Previous works attempt to address these challenges by training VLMs with behavior cloning. Agents have been trained to parse GUIs~\cite{cheng2024seeclick, gou2024navigating,lu2024omniparser} and make plans based on screen captures~\cite{hong2024cogagent,cheng2024seeclick}. These approaches, however, heavily depend on large amounts of well-annotated GUI data and real-world trajectory data, which are extremely expensive and labor-intensive to obtain. Moreover, the alignment issue of LLMs~\cite{shen2023large} also occurs with the VLMs and vision agents. Hence, undesired actions in a trajectory often lead to failures in completing the agent's objective.

In this paper, we present a {\underline{ste}}p {\underline{ve}}rification pipeline coined STEVE, a new approach that automatically verifies the correctness of agent actions with existing large VLMs, for providing dense, stepwise reward signals to agent training. Compared with the traditional reinforcement learning (RL) setting, STEVE does not require carefully handcrafted reward functions in a computer environment~\cite{xie2024osworld, bonatti2024windows}, enabling us to largely upscale the number of tasks and train better computer-use agents in desktop environments. 

Our approach consists of three major steps. First, we collect a large dataset of web pages and desktop screenshots to train a VLM specialized in UI grounding. The model is fine-tuned into a computer-use agent with limited trajectory data in a supervised learning way. Then, we deploy the agent in a live Windows environment and collect a large number of trajectories. As Fig.~\ref{fig:framework} shows, we leverage GPT-4o as a step verifier to evaluate each action and obtain an upsized dataset with stepwise rewards. Last, we optimize the suboptimal agents with the Kahneman-Tversky Optimization~\cite{ethayarajh2024kto} on the collected step-verified trajectories. 

Extensive experiments were conducted to compare our trained agents with supervised finetuning (SFT) agents.
The results show that our agents can make full use of the data and scale more effectively than SFT with increased training tasks and trajectories. Besides, when jointly training the models with UI-grounding data and agent task data, SFT causes a severe degradation in UI localization precision, while 
our STEVE-trained agent is able to perfectly inherit the capability from the UI-grounding model. 

Our main contributions are summarized as follows:

\begin{itemize}
    \item A powerful GUI-grounding VLM: Our model sets a new state of the art
    on several UI localization benchmarks, especially a new record on the challenging WindowsAgentArena live environment.
    \item The scalable step verification pipeline STEVE: we carefully design it to automatically upsize the agent instruction set for producing a large trajectory dataset with GPT-verified stepwise rewards for agent training.
    \item KTO optimization to utilize both the positive and negative actions from the step verification pipeline for computer-use agent training.
    The experiments show that our trained agents effectively leverage both positive and negative samples in the trajectory data and avoid degrading the agent's UI localization ability. 
\end{itemize}

\section{Related works}


\noindent\textbf{Screen UI understanding.} 
Recent advances in GUI agents leverage large vision language models (VLMs) for interacting with user interfaces. Qwen2-VL~\cite{wang2024qwen2} introduces GUI data to train a general VLM to learn UI understanding. UGround~\cite{gou2024navigating} and Ferret UI~\cite{li2024ferret} introduce a specialist visual grounding model that significantly improves GUI agents in mapping textual instructions to precise GUI elements. OmniParser~\cite{lu2024omniparser} offers a screen parsing tool that extracts structured elements from UI screenshots, enhancing GPT-4V's action prediction on various platforms, without requiring additional input beyond screenshots.

\vspace{1mm}
\noindent Recent datasets substantially advance UI interaction research. RICO~\cite{deka2017rico} supports UI design and interaction modeling.
WebUI~\cite{wu2023webui} provides web pages for visual UI understanding.
AITW~\cite{rawles2024androidinthewild} focuses on Android device control with multi-step tasks.
Mind2Web~\cite{deng2024mind2web} targets generalist agents for complex tasks on real websites, whereas GUICourse~\cite{chen2024guicourse} enhances the VLM's abilities in GUI interaction with various GUI conversation data. These resources push the boundaries of web and mobile UI automation.

\vspace{1mm}
\noindent\textbf{Computer-use agents.} 
On the other hand, recent multimodal models spark significant progress in GUI and web automation. SeeClick~\cite{cheng2024seeclick} and ScreenAgent~\cite{niu2024screenagent} leverage visual inputs for task automation; the former focuses on GUI grounding pre-training and the latter on building agents that interact with real computer screens. OmniAct~\cite{kapoor2024omniact} extends these efforts with a benchmark for generating executable scripts based on visually-grounded natural language tasks. CogAgent~\cite{hong2024cogagent} pre-trains models with a large amount of web and desktop data for 
screen UI localization. 

\vspace{1mm}
\noindent Recent works such as SEEACT~\cite{zheng2024gpt}, UFO~\cite{zhang2024ufo}, and Agent S~\cite{agashe2024agent} tackle GUI task automation by designing an agent workflow that integrates grounding, control, and planning. SEEACT~\cite{zheng2024gpt} focuses on visually-grounded web interaction, UFO~\cite{zhang2024ufo} on seamless control across Windows applications, and Agent S~\cite{agashe2024agent} on hierarchical planning for multi-step, long horizon complex task execution.
%
\noindent OSWorld~\cite{xie2024osworld} and WindowsAgentArena (WAA)~\cite{bonatti2024windows} introduce scalable, real computer environments for evaluating multimodal agents. OSWorld spans multiple operating systems,
while WAA~\cite{bonatti2024windows} focuses on Windows OS, both offering a dynamic real-world environment to agent evaluation.

\vspace{1mm}
\noindent\textbf{Reinforcement learning for LLMs.}
Reinforcement learning plays a key role in aligning LLMs with human preferences. Proximal Policy Optimization (PPO)~\cite{schulman2017proximal} is commonly used for training LLMs with human feedback (RLHF) due to its balance of stability and performance, but its complexity and cost have led to alternatives.
Direct Preference Optimization (DPO)~\cite{rafailov2024direct} simplifies RLHF by removing the need for reward modeling. Recently, RLOO~\cite{ahmadian2024back} has shown that less computationally expensive approaches can outperform PPO, highlighting a trend toward more efficient RL for LLM alignment. KTO~\cite{ethayarajh2024kto} incorporates human biases from prospect theory for better alignment. In this work, we discuss how stepwise environmental feedback can help align computer agents with human preferences.

\noindent \textbf{Step verification for LLMs.} 
Recent work emphasizes the importance of verifying every reasoning step in long-chain tasks to improve the performance of LLMs. Process supervision, as shown in ``Let’s Verify Step by Step~\cite{lightman2023let},'' is proved to be more effective than outcome-based feedback, especially for complex datasets like MATH~\cite{hendrycks2021measuring}.
MathShepherd~\cite{wang2024math} further automates step-by-step verification and reinforcement using process-wise supervision, largely enhancing LLM performance without heavy reliance on human annotations. Step-DPO~\cite{lai2024step} builds on this by optimizing individual steps instead of the final answers, improving accuracy in mathematical reasoning with fewer data and training steps. These approaches collectively demonstrate the critical role of step-level verification and inspire us to design stepwise supervisions to train computer-use agents.

\section{Method}
\label{sec:method}

\begin{figure}[t]
    \resizebox{\linewidth}{!}{
        \begin{tabular}{lccc}
        \hline
        \textbf{Dataset} & \textbf{Annotation} & \textbf{Num. Image} & \textbf{Num. elements} \\
        \hline
        WebUI~\cite{wu2023webui} & DOM, OCR & 180K & 1M \\
        Seeclick~\cite{cheng2024seeclick}& DOM & 10K & 150K \\ 
        AITW~\cite{rawles2024androidinthewild} & Caption & 15K & - \\
        Allava~\cite{chen2024allava} & general QA & 50K & - \\
        Windows OS & A11y, GPT-4o & 10K & 80K \\
        \hline
        \end{tabular}
    }
    \caption{Datasets we collected for UI-grounding model training, including open-source datasets and an additional private Windows OS dataset created by ourselves to enhance the model's performance on Windows.
    }
    \label{tab:dataset}
\end{figure}

In this section, we present STEVE, the step verification training pipeline for our computer-use agent. 
Our approach starts from a UI-grounding vision language model and then integrates agent task training to enable the model to solve multi-step tasks in a desktop environment.

\begin{figure*}[th!]
\centering
    \includegraphics[width=\linewidth]{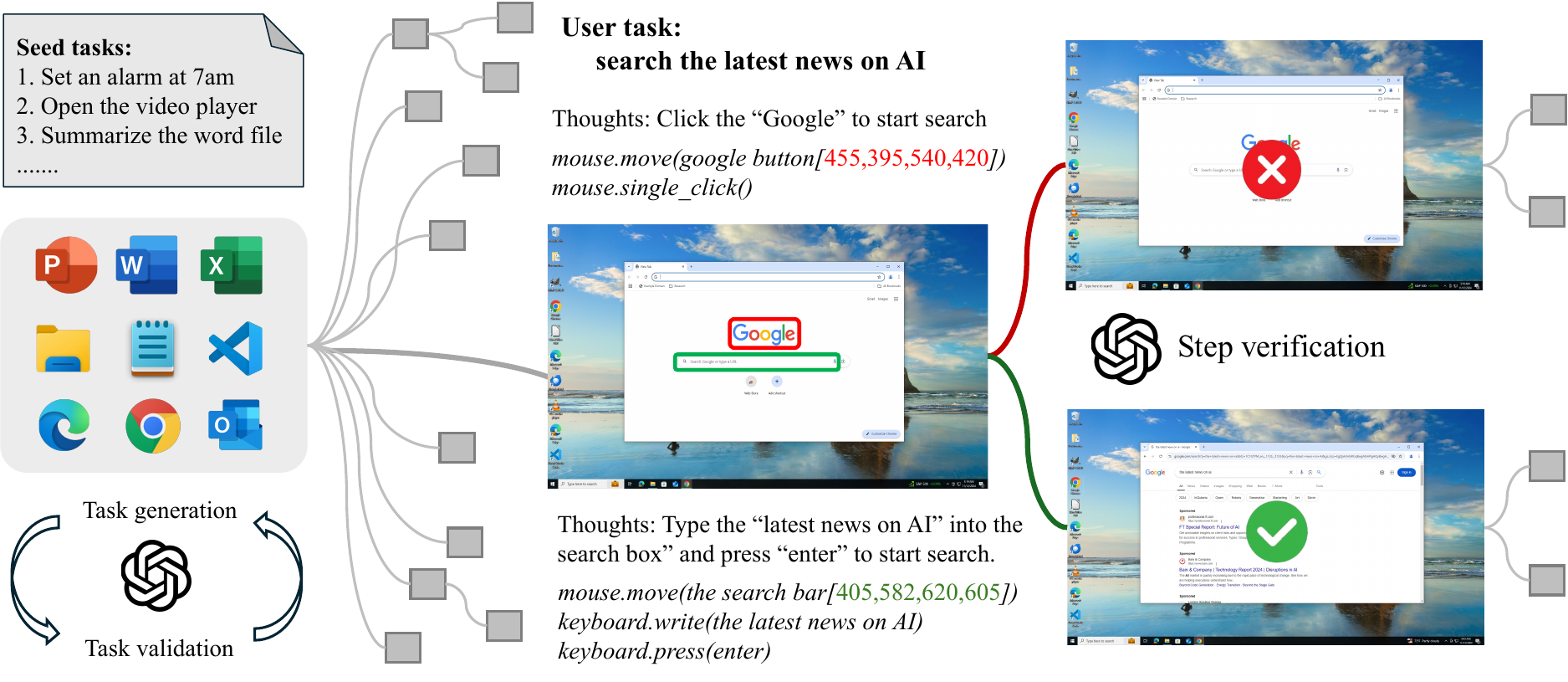}
    \caption{
       Overview of STEVE, the step verification pipeline.
        We first create a large number of feasible tasks from the seed tasks to scale up the quality and diversity of agent tasks. Then we deploy our computer-use agent in desktop environments to sample trajectory data. A GPT-4o judge is used to verify the quality of each step in the trajectory, resulting in a large process reward dataset for agent training.
    }
\label{fig:framework}
\end{figure*}

\subsection{UI-grounding Model}

A robust UI understanding and grounding model is crucial for building an effective computer-use agent. To train our UI-grounding model, we collected a large amount of web and desktop screenshot data.

\noindent \textbf{Web data.}
For web data, we parsed the DOM (Document Object Model) of numerous web pages~\cite{wu2023webui} to first extract all text-based UI elements and their corresponding bounding boxes. We then further refine these text elements and remove noise that may have been introduced during the DOM parsing. Also, we applied an OCR model~\cite{du2020pp} to validate the extracted UI element.

\vspace{1mm}
\noindent \textbf{Desktop screenshot.}
For desktop screenshot data, we set up a Windows virtual machine (VM) and leveraged an existing OmniParser~\cite{lu2024omniparser} to perform tasks within the VM environment. During the task execution, we captured screenshots and gathered the associated accessibility tree (A11y Tree) data. Also, we designed specific rules to filter out noisy results from the A11y Tree, thereby enabling us to collect 10k desktop images and 80k UI elements. Additionally, we incorporated a portion of publicly available AITW data to further augment our dataset.

\vspace{1mm}
\noindent \textbf{Screenshot captioning.}
Beyond UI-grounding data, we collected 30k high-quality captions to further enrich the dataset and facilitate the understanding of the screen captures and UI elements of our VLMs during training.

In summary, based on the aforementioned data, we trained a large vision-language model capable of accurately grounding UI elements in 1080p resolution screenshots. Compared to previous methods, our approach demonstrated significant improvements across several benchmarks, including ScreenSpot~\cite{cheng2024seeclick}, AITW~\cite{rawles2024androidinthewild}, and Mind2Web~\cite{deng2024mind2web}. We further integrated this grounding model into an agent framework, inspired by the WinAgentArena~\cite{bonatti2024windows} architecture, where GPT-4o was employed as the planner. The planner model is responsible for understanding the user instructions and delegating commands to the grounding model. This agent framework achieved a 22\% task success rate on the challenging WinAgentArena, surpassing the previous state-of-the-art results~\cite{lu2024omniparser} on this benchmark.

\tcbset{
  colback=gray!5!white, 
  colframe=black,       
  width=\linewidth,     
  boxrule=1pt,          
  arc=4mm,              
  left=5pt,             
  right=5pt,            
  top=5pt,              
  bottom=5pt,           
}

\subsection{Computer-use Agent Finetuning} 
\label{sec:agent_sft}
We found that it is non-trivial to finetune a UI-grounding model into a high-performing agent. We present the prompts we employed for both the UI-grounding model and the agent model as follows:

\begin{tcolorbox}[title=UI-grounding Model Prompt Example]
\footnotesize 
User query: Provide the bounding box for the {\texttt{<target>}}.\\
Answer: \texttt{<target>}\texttt{<box>}(x0,y0),(x1,y1)\texttt{</box>}
\normalsize
\end{tcolorbox}

\begin{tcolorbox}[title=Computer-use Agent Prompt Example]
\footnotesize 
User query: You are a helpful assistant. You have the actions: \\
\# \texttt{Action definitions}\\
The current screen image: \texttt{<image>}. The user task: \texttt{<task>}.\\
The history information: \texttt{<history>}.\\
Answer: \texttt{<chain of thought analysis>}. \\
Action: mouse.move({\texttt{<target>}\texttt{<box>}(x0,y0),(x1,y1)}\texttt{</box>})
\normalsize
\label{text:prompt}
\end{tcolorbox}

There is a significant distributional discrepancy between the training data of the UI-grounding model and that of the agent model. When we attempted to directly finetune 
the UI-grounding model on agent data, it led to a severe degradation in the model's UI localization capabilities.

To mitigate this issue, we explored two potential approaches: (i) freezing the weights of the UI-grounding model and using a LoRA adapter~\cite{hu2021lora} for finetuning, and (ii) mixing UI-grounding data with agent data during the finetuning process. However, neither approach was sufficient to address the degradation problem. A detailed analysis of this issue, along with a comparison of various methods, is presented in Section~\ref{sec:ui_acc}.

\subsection{Step Verifier for Trajectory Evaluation}

RL environments typically provide agents with sparse reward signals only at the end of tasks, thus leading to extremely inefficient exploration.  Behavior cloning, on the other hand, requires expensive trajectory data with step-wise annotations. To circumvent the shortcomings, we propose a step verification mechanism that evaluates the quality of each action taken by the agent within a task trajectory.

\vspace{1mm}
\noindent \textbf{Visual feedbacks from the environment.}
Different from conventional step-verification methods for improving the math and reasoning ability of LLMs~\cite{lightman2023let, wang2024math}, as illustrated in Fig.~\ref{fig:framework}, the incorrect actions, such as invalid or erroneous clicks, can be easily distinguished from the correct ones by comparing the screens before and after the agent's action. This direct feedback mechanism significantly simplifies the evaluation of step-wise actions within a trajectory. 

We found that the general visual capabilities of large powerful VLMs, such as GPT-4o, allow for highly accurate evaluations, which aligned well with human judges. The data format for this evaluation is as follows:
\begin{equation}
y_t = V(x_t, \{r_t, a_t\}, x_{t+1}),
\end{equation}
where $r_t$ is the chain-of-thought reasoning generated by the agent to address the current step of the user task, and $a_t$ is the action proposed by the agent model for the current screenshot $x_t$. Here, $V$ is a large VLM judge, such as GPT-4o, and $y_t$ is a binary annotation for $a_t$. An action is verified as beneficial, if both the reasoning is correct and the action is correctly executed, which results in the expected transition from screen $x_t$ to screen $x_{t+1}$. 

Therefore we assign positive or negative annotations to each action. This step verifier provides valuable feedback for the agent's learning process, enabling more efficient and targeted improvements in performance. Examples of agent trajectories and step verification results can be found in the appendix.

\vspace{1mm}
\noindent \textbf{Task instruction scaling up.}
The step verification mechanism we propose does not require the design of complex reward functions for each task, unlike the WinAgentArena~\cite{bonatti2024windows} and OSWorld~\cite{xie2024osworld}, where about only 200 tasks are designed with elaborated reward signal. Instead, it relies on a powerful VLM as the evaluator, which allows us to scale up the task instructions. We start with a batch of seed tasks~\cite{alpaca} and use GPT4 to generate new tasks by editing the seed tasks and creating similar tasks. 

\vspace{1mm}
\noindent \textbf{Task feasibility.}
It is important to ensure the feasibility of the tasks generated since infeasible tasks contribute little to the trajectory sampling. To address the problem, we provide the GPT4 with real-world files and documents and prompt the model to generate feasible instructions from a batch of seed tasks. We employ the GPT-o1 model to verify the feasibility of the tasks. Ultimately, we synthesized over 4,000 tasks in the Windows environment, covering various scenarios such as OS settings, file explorer, Windows app, and website browsing. Detailed examples of these tasks can be found in the appendix.

\subsection{KTO Training with Stepwise Rewards}
The previous works~\cite{gou2024navigating, agashe2024agent, lu2024omniparser, zhang2024ufo} usually design two-stage computer agent systems, where a preprocessing model is used to extract all the GUIs into structured elements and then adopt a planning model such as GPT-4o to make multi-step planning and decisions. In contrast, we aim to train a single agent model that is capable of both low-level UI perception and high-level decision and multi-step planning.

We leveraged the synthesized tasks in parallel Windows environments~\cite{bonatti2024windows} to enable the agent to execute and log screenshots and actions during task execution. Afterward, we used the GPT-4o verifier to annotate each step in the trajectory, resulting in a large-scale dataset with stepwise annotations. In the following, we describe how we utilized this data to train a more effective computer-use agent.

\vspace{1mm}
\noindent \textbf{Iterative finetuning.} A straightforward approach is iterative finetuning~\cite{dong2023raft}. As the agent produces trajectory data in an environment, the positive samples verified as successful are iteratively selected to UI-grounding the agent model. Yet, these approaches are data inefficient, as they neglect the negative samples, which in fact can also contribute.

\vspace{1mm}
\noindent \textbf{Direct Preference Optimization.} \
DPO~\cite{rafailov2024direct} requires paired positive and negative samples for training. This approach has been shown to be highly reliable in LLM finetuning. However, due to the complexity of the machine states and task trajectories, it is difficult to collect paired positive and negative data. On the other hand, we can easily collect a full trajectory and evaluate each step of it.  

\vspace{1mm}
\noindent \textbf{Kahneman \& Tversky Optimization.} \
The limitations of iterative finetuning and DPO can be effectively addressed by KTO~\cite{ethayarajh2024kto}, which offers various advantages:
(i) 
KTO can be trained with unpaired positive and negative samples, eliminating the need for paired data, which takes huge human efforts to obtain in the desktop environment.
(ii) 
The agent’s poor performance in the early stage leads to a significant imbalance between positive and negative samples. KTO effectively handles this data imbalance for a more stable optimization.
(iii) 
KTO needs only binary reward scores (+1/-1) for training, 
promoting the training process with higher stability and robustness.

\vspace{1mm}
We adopt the vanilla KTO loss for training:
\begin{equation}
    L_{\text{KTO}}(\pi_{\theta}, \pi_{ref}) = \mathbb{E}_{x,y\sim D}[\lambda_y - v(x, y)]
\end{equation} 
where
\begin{align}
    r_\theta(x, y) &= log\frac{\pi_{\theta}(y|x)}{\pi_{ref}(y|x)} \\
    z_0 &= \text{KL}(\pi_\theta(y\prime|x || \pi_{ref}(y\prime |x)) \label{eq:z}\\
    v(x, y) &= 
    \begin{cases} 
    \lambda_D \sigma\left( \beta \left( r_\theta(x, y) - z_0 \right) \right) & \text{if } y \sim y_{\text{desirable}} | x \\
    \lambda_U \sigma\left( \beta \left( z_0 - r_\theta(x, y) \right) \right) & \text{if } y \sim y_{\text{undesirable}} | x.
    \end{cases}
\end{align}
The $\lambda_D$ and $\lambda_U$ are hyperparameters for the desired and undesired data, respectively. $\lambda_y$ denotes $\lambda_D$, when $y$ is desirable, otherwise $\lambda_U$. Eq.~\eqref{eq:z} denotes a biased estimation of the KL divergence~\cite{ethayarajh2024kto}.

\vspace{1mm}
\noindent \textbf{KTO initialization and training} We use our UI-grounding model with the GPT-4o planner to collect trajectories and finetune the grounding model to a reference policy model. Then, we perform our KTO training process by repeatedly sampling trajectories in the live Windows environment and gradually increasing the number of trajectories to 4,820. During the KTO training stage, to optimize the memory usage and performance, we use two separate LoRA~\cite{hu2021lora} adapters as the reference model and the actor model, thereby largely reducing the memory overhead. 


\vspace{1mm}
\noindent \textbf{Multi-round KTO.} Since we sample the trajectories using a single VLM agent, the policy $\pi_{\theta}$ is fixed and the negative actions may fall into a narrow distribution. To mitigate the problem, we leverage a multi-round trajectory collection and KTO training. By conducting multiple rounds of trajectory sampling, we expose the agent to a variety of scenarios, enabling us to more efficiently explore a broader action space. This increased diversity also helps prevent overfitting to a limited set of non-optimal actions and improves the generalization of the KTO optimization.

\section{Experiments}
We evaluate the GUI localization capability of our grounding model on the ScreenSpot~\cite{cheng2024seeclick} and AITW benchmarks and our agent model on the WindowsAgentArena~\cite{bonatti2024windows} and Mind2Web~\cite{deng2024mind2web} benchmarks. We finetune our models from the Qwen2-VL~\cite{wang2024qwen2} model. All the prompts, question templates, and training details for the grounding and agent models can be found in the appendix.

\begin{table*}[t]
    \centering
    \resizebox{0.7\linewidth}{!}{
    \begin{tabular}{ccccccccc}
         \hline
         \multirow{2}{*}{Method} & \multirow{2}{*}{Size} & \multicolumn{2}{c}{Mobile} & \multicolumn{2}{c}{Desktop} & \multicolumn{2}{c}{Web} & \multirow{2}{*}{Overall} \\
        & & Text & Widget & Text & Widget & Text & Widget & \\
         \hline
         Qwen-VL & 9.6B & 9.5 & 4.8& 5.7& 5.0& 3.5& 2.4& 5.2\\
         Fuyu & 8B & 41.0 & 1.3 & 33.0 & 3.6 & 33.9 & 4.4 & 19.5 \\
         CogAgent~\cite{hong2024cogagent} & 18B & 67.0 & 24.0 & 74.2 & 20.0 & 70.4 & 28.6 & 47.4 \\
         Seeclick~\cite{cheng2024seeclick} & 9.6B &78.0 & 52.0 & 72.2 & 30.0 & 55.7& 32.5& 53.4 \\
         Qwen2-VL~\cite{wang2024qwen2} & 7B & 75.5 & 60.7 & 76.3 & 54.3 & 35.2 & 25.7 & 55.3 \\ 
         OmniParser~\cite{lu2024omniparser} & GPT-4o & 93.9 & 57.0 & 91.3 & 63.6 & 81.3 & 51.0 & 73.0 \\
         UGround~\cite{gou2024navigating} & 7B & 82.8 & 60.3 & 82.5 & 63.6 & 80.4 & 70.4 & 73.3 \\  
         \hline
         Ours & 7B &  88.6 & \textbf{81.2} & 88.1 & \textbf{78.6} & 78.2 & \textbf{76.2} & 82.2  \\
          Ours$\dagger$ & 7B & \textbf{94.9} & 80.0 & \textbf{94.3} & 70.7 & \textbf{87.0} & 70.4 & \textbf{84.0} \\
         \hline
    \end{tabular}
    }
    \caption{The performance on the GUI localization benchmark ScreenSpot~\cite{cheng2024seeclick}. $\dagger$ indicates the self-plan evaluation~\cite{gou2024navigating} using GPT-4o generated reference expressions as queries to the model.}
    \label{tab:screenspot}
\end{table*}

\begin{table}[t]
    \centering
    \resizebox{\linewidth}{!}{
    \begin{tabular}{ccccccc}
       \hline
       Method  & GoogleApp & Install & WebShop & General & Single & Overall \\
       \hline
       GPT3.5(few-shot) & 10.5 & 4.4 & 8.4 & 5.9 & 9.3  & 7.7 \\
       LLaMA2  & 31.0 & 35.2 & 19.9 & 28.6 & 27.4 & 28.4 \\
       CogAgent~\cite{hong2024cogagent}  & 50.8 & 57.1 & 49.7 & 47.6 & 43.4 & 49.7\\
       SeeClick~\cite{cheng2024seeclick} & 57.7 & 64.5 & 57.3 & 56.0 & 63.6 & 59.8 \\
       GUICourse~\cite{chen2024guicourse}  & 70.3 & 61.2 & \textbf{71.6} & - &66.1& 67.3 \\
       \hline
       Ours  & \textbf{70.7} & \textbf{75.4}& 69.7 & \textbf{66.4} & \textbf{78.9} & \textbf{72.2} \\
       \hline
    \end{tabular}
    }
    \caption{Results on the AITW benchmark. We use the same test split as SeeClick~\cite{cheng2024seeclick}. The step-level action accuracy is reported.}
    \label{tab:aitw}
\end{table}

\begin{table}[t]
    \centering
    \resizebox{\linewidth}{!}{
    \begin{tabular}{cccccccc}
         \hline
         \multirow{2}{*}{Method} & \multicolumn{2}{c}{Task} & \multicolumn{2}{c}{Website} &  \multicolumn{2}{c}{Domain} & \multirow{2}{*}{Overall}  \\
         & Step SR & Elem Acc & Step SR & Elem Acc & Step SR & Elem Acc & \\
         \hline
         CogAgent & 17.6 & 22.4 & 13.4 & 18.4 & 15.5 & 20.6 & 15.5 \\
         Qwen-VL & 14.9 & 14.1 & 12.1 & 13.2 & 9.7& 14.1 & 12.2 \\
         SeeClick & 25.5 & 28.3 & 16.4 & 21.4 & 20.8 & 23.2 & 20.9 \\
         OmniParser & 39.4 & 42.4 & 36.5 & 41.0 & \textbf{42.0} & 45.5 & 39.3 \\ 
         \hline
         Ours & \textbf{40.0} & \textbf{46.2} & \textbf{37.7} & \textbf{44.4} & 41.2 & \textbf{46.0} & \textbf{39.6} \\
         \hline
    \end{tabular}
    }
    \caption{Results on the Mind2Web benchmark.}
    \label{tab:mind2web}
\end{table}

\begin{table*}[t!]
    \centering
    \resizebox{\textwidth}{!}{
    \begin{tabular}{cccccccccc}
    \hline
    Method & Size & A11y & Office & Web Browser & Windows System & Coding & Media  Video & Windows Utils & Overall \\
    \hline
    OmniParser & GPT-4o & \checkmark  & 0.0 & 13.7 & 29.2 & 0.0 & 10.3 & 0.0 & 8.6 \\
    NAVI & GPT-4o  & \checkmark & 0.0 & 20.0 & 29.2 & 9.1 & 25.3 & 0.0 & 13.3 \\
    OmniParser & GPT-4V-1106 &  & 2.3 & 23.6 & 20.8 & 8.3 & 20.0 & 0.0 & 12.5 \\
    Agent S  & GPT-4o & \checkmark & 0.0 & 13.3 & 45.8 & 29.2& 19.1 & \textbf{22.2} & 18.2 \\
    OmniParser & GPT-4V-1106 & \checkmark & 0.0 & 27.3 & 33.3 & 27.3 & \textbf{30.3} & 8.3 & 19.5 \\
    \hline
    Ours-SFT & 7B & & 2.3 & 21.0 & 20.8 & 0.0 & 0.0 & 0.0 & 7.1 \\
    Ours-KTO & 7B & & 2.3 & 36.8 & 37.5 & 16.6 & 9.5 & 0.0 & 14.2 \\
     Ours-G & GPT-4o & & \textbf{4.6} & \textbf{52.4} & \textbf{45.8} & 20.8 & 11.8 & 16.7 & \textbf{23.0} \\
    \hline
    \end{tabular}
    }
    \caption{Performance on the WinAgentArena benchmark. ``Our-G'' denotes our UI-Grounding model with the GPT-4o planner.}
    \label{tab:waa}
    
\end{table*}

\subsection{GUI Grounding Evaluation}
\label{exp:ui_ground}

\noindent \textbf{ScreenSpot.} We first evaluate the performance of our UI-Grounding model on the ScreenSpot~\cite{cheng2024seeclick} benchmark, a dataset that contains more than 1,000 queries about GUIs in static screenshots. The dataset covers website, desktop, and mobile domains with text and widget UI types in each domain. The task is to correctly locate the position of the UI according to the language instruction. 
We represent the performance in Tab.~\ref{tab:screenspot} that our UI-grounding model performs the previous state-of-the-art methods by $8.9\%$ on the GUI-Grounding task. The precise UI localization ability plays an important role in the later stage of training a powerful computer agent. With GPT-4o refined instruction to the UI element, our model achieves a score of $84.0\%$, which is more than 10 points beyond the best SOTA method.

\noindent \textbf{AITW.} Android in the wild~\cite{rawles2024androidinthewild} provides a large mobile dataset for training and evaluating mobile agents. The actions include tapping, texting, scrolling, and button pressing on an Android device. We take 200K training screenshots from the train split to finetune the grounding model for the downstream application. To align with previous works~\cite{cheng2024seeclick, chen2024guicourse}, we take the same test split to evaluate the performance of our model. The step action success rate is reported.
Tab.~\ref{tab:aitw} demonstrates that our UI-Grounding model can be easily finetuned for downstream tasks and achieve $4.9\%$ performance gain over UGround~\cite{gou2024navigating}, which is the previous best vision language model on the benchmark. 

\vspace{1mm}
\noindent \textbf{Multi-Modal Mind2Web.} We also evaluate our UI-Grounding model on the Multi-Modal Mind2Web~\cite{deng2024mind2web} to examine the performance on realistic web browsing tasks. The benchmark consists of 1,013 real tasks from Cross-Website, Cross-Domain, and Cross-Task categories respectively. Each task in Mind2Web is described with a high-level user instruction and the agent has to select from three available actions: clicking, typing, and selecting. We report stepwise success rate and element accuracy on the benchmark. Following the paper~\cite{gou2024navigating}, our UI-Grounding model uses a GPT-4o planner for high-level task planning and uses the reference expression created from GPT-4o to localize the position of the target UI. See Tab.~\ref{tab:mind2web} Our method outperforms SeeClick by $18.7\%$ in stepwise success rate and OmniParser by $0.3\%$, while being 20 times faster.

\subsection{Computer-use Agent Evaluation}
Next, we present evaluations on the live Windows OS benchmark, WinAgentArena~\cite{bonatti2024windows}, a comprehensive benchmark to evaluate computer-use agents in Windows OS. The environment provides 154 tasks from the office, web browsing, Windows system, coding, media, and Windows apps domains. Each task comes with a handcrafted reward function to measure whether the task is complete or not. It takes an average number of $7$ steps to complete a task. 

Tab.~\ref{tab:waa} presents a comparison of our method with OmniParser and Agent S. Both of the methods adopt GPT-4o as their task planner. Our 7B UI-Grounding model with the GPT-4o planner outperforms the other approaches and sets a new state of the art on the challenging WinAgentArena benchmark. Besides, we show the performance of our 7B agent model trained with SFT and KTO. This is the first work that achieves a record with a 7B model.

\subsection{Component-wise Analysis.}

\begin{figure}[ht]
\centering
    \includegraphics[width=0.85\linewidth]{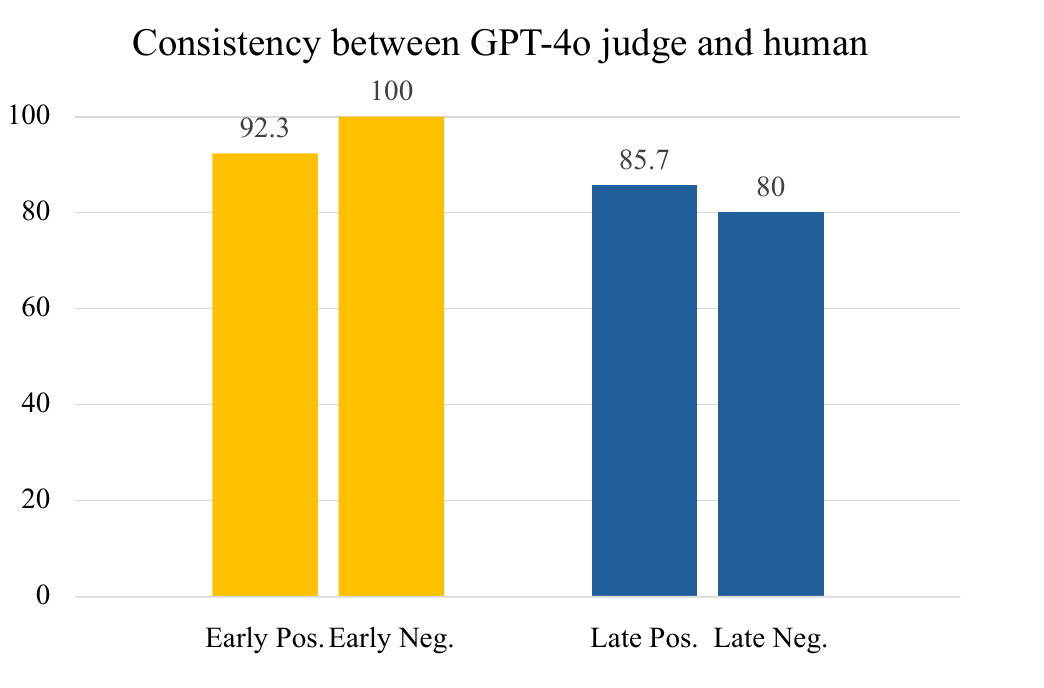}
    \caption{
    Percentage consistency between human judges and the GPT-4o step verifier. We split all the positive and negative actions into early (step ID $\le 7$) and late (step ID $> 7$) groups, resulting in four bars in the figure. For example, $92.3\%$ for the Early Pos. bar means the GPT-4o judge agrees with humans for $92.3\%$ of the early positive actions.
}
\label{fig:consist}
\end{figure}

\noindent \textbf{Human consistency.} It is necessary to validate the consistency between a GPT-4o verifier and human judges. We randomly sample a subset of trajectories and manually annotate each step. To properly assess the consistency, we categorize the actions into early and late phases based on the order of occurrence within a trajectory.  In order to assess the consistency more rigorously, we divided the actions into two phases: early and late, based on their order of occurrence within a trajectory. As illustrated in Fig.~\ref{fig:consist}, the GPT-4o verifier demonstrates a high degree of alignment with human judgments during the early phase of the trajectory steps. However, as the task progresses into the late phase, the verifier's precision decreases. This reduction in accuracy can be attributed to the increased complexity of the later steps, where the evaluation of actions becomes more challenging due to dependencies on both the current step and preceding actions within the trajectory.

\begin{figure*}[htbp]
    \centering
    \begin{subfigure}[b]{0.3\textwidth}
        \centering
        \includegraphics[width=\textwidth]{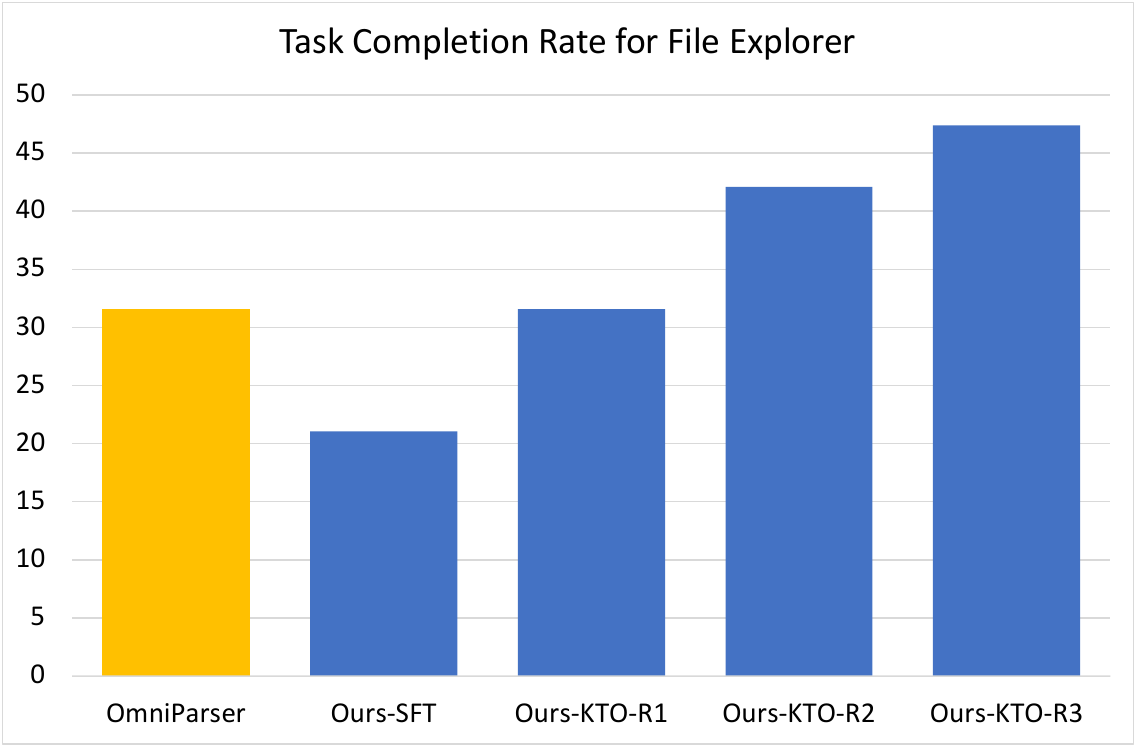} 
        \caption{Task success rate on the File Explorer split.}
        \label{fig:subfig1}
    \end{subfigure}
    \hfill
    \begin{subfigure}[b]{0.3\textwidth}
        \centering
        \includegraphics[width=\textwidth]{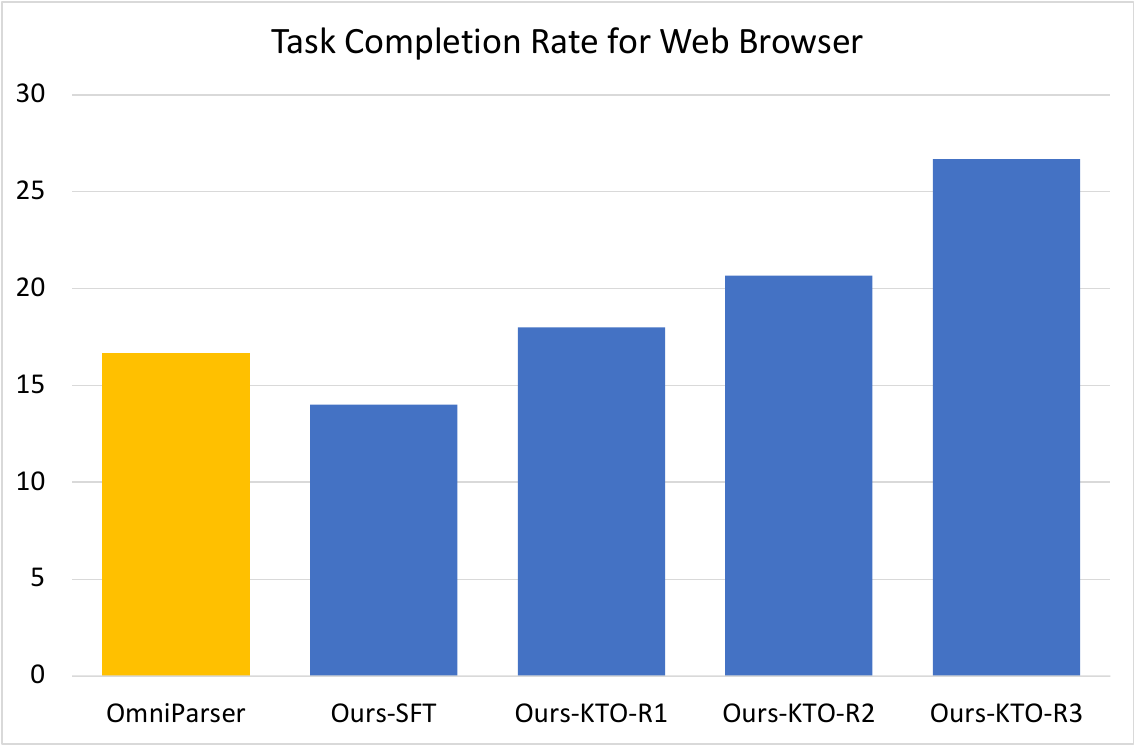} 
        \caption{Task success rate on the Web Browser split.}
        \label{fig:subfig2}
    \end{subfigure}
    \hfill
    \begin{subfigure}[b]{0.3\textwidth}
        \centering
        \includegraphics[width=\textwidth]{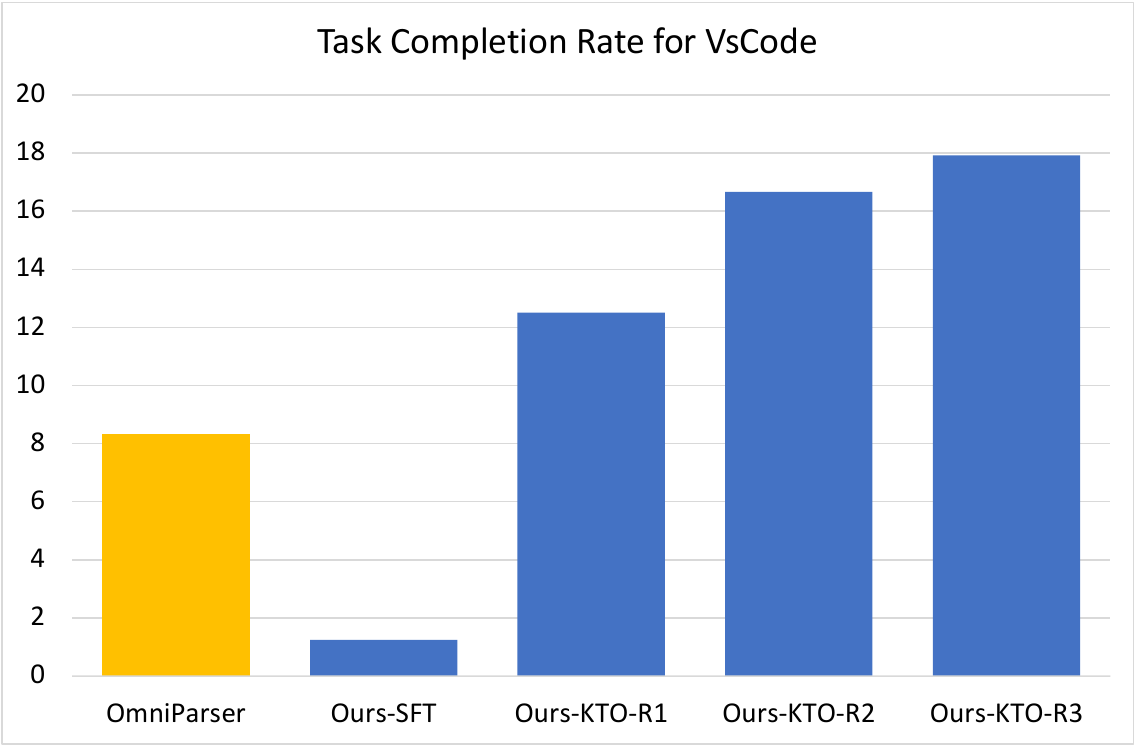} 
        \caption{Task success rate on the VsCode split.}
        \label{fig:subfig3}
    \end{subfigure}

    \caption{We show an ablation study of OmniParser, the SFT agent, and three KTO agents at three iterative rounds (SFT, R1, R2, and R3). The results are evaluated on three distinct task domains from the WinAgentArena benchmark. Yellow bars in the figures indicate that GPT-4o is employed as the task planner. The reported outcomes represent the average performance over five experimental runs.}
    \label{fig:ablation_split}
\end{figure*}



\vspace{1mm}
\label{sec:ui_acc}
\noindent \textbf{SFT or KTO for precise UI localization?} We observe that supervised finetuning a UI-Grounding model with agent planning data causes significant degradation to the UI localization performance, especially on high-resolution screenshots. The situation is even worse when the trained agent model works with the agent prompt template, as defined in ~\ref{text:prompt}. To better understand this degradation, we explore various finetuning strategies, including standard SFT, LoRA SFT, and mixed data training, and evaluate their impact on UI localization performance.

See Tab.~\ref{tab:ui_sft}, we conduct comprehensive experiments to evaluate the different training strategies. We take 15K verified actions for the ablation study, with an equal number of positive and negative action steps. For the SFT setting, only the positive verified data is used for training, while for KTO model, both are used. For the mixed data training setting, we augment the agent planning data by incorporating UI-Grounding data and double the size of the training set, mitigating the effect of domain shift between datasets. All numerical results in Tab.~\ref{tab:ui_sft} are measured using the UI-Grounding prompt template.

\begin{figure}[h]
    \centering
    \includegraphics[width=\linewidth]{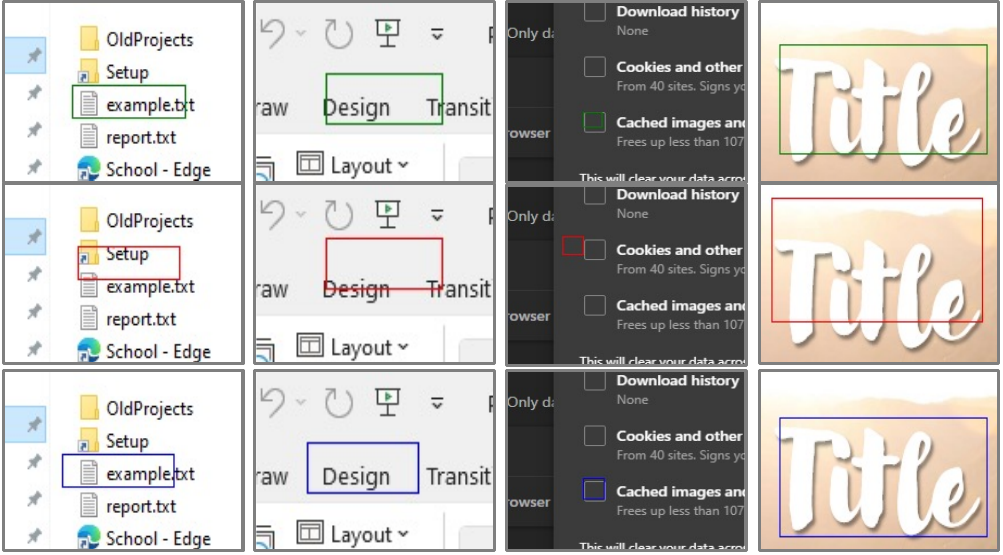}
    \caption{Zoom in visualization of UI localization performance of different models on four target GUIs: example.txt, Design tab, Cached image check box, and Title of PPT slide (left to right). The UI-Grounding Model's performance is shown in green (top row), the SFT-trained agent in red (middle row), and the KTO-trained agent in blue (bottom row).}
    \label{fig:bbox}
\end{figure}

Our results indicate that agents trained using SFT exhibit poor performance in recognizing small UI elements. We categorize the UI elements in the ScreenSpot benchmark into three groups based on their size: small (elements with a maximum side of less than 50 pixels), medium (less than 100 pixels), and large (100 pixels or more). As illustrated in Tab.~\ref{fig:ui_acc}, SFT training results in a performance decline of $6.1$\% for small UI elements and $2.0$\% for medium-sized elements, leading to an overall performance drop of $1.7$\%. In contrast, the KTO model shows a smaller reduction of $2.0$\% in performance for small UI elements, while improving by $2.0$\% for medium-sized elements, resulting in a slight overall performance increase of $0.3$\%.

We visualize the UI Grounding results of the UI Grounding model, SFT-trained agent, and KTO-trained agent in Fig.~\ref{fig:bbox}. The models' predictions on four target GUIs are shown with green (UI Grounding), red (SFT), and blue (KTO) bounding boxes. We show that our KTO training allows the agent model to inherit and even surpass the UI localization precision of the UI-Grounding model. We posture that the KTO agent, optimized to avoid invalid and erroneous clicks, learns a better embedding and predicts tight bounding boxes to the target GUIs.

\begin{table}[t!]
 \centering
    \resizebox{\linewidth}{!}{
    \begin{tabular}{cccccc}
    \hline
    Models & Data & Small & Middle & Large & Overall \\
    \hline
    Base UI model&  & \text{67.3} & 74.6 & 84.5 & 82.2 \\
    \hline
    SFT  & agent & 61.2(-6.1) & 72.6(-2.0) & 84.4 & 80.5(-1.7) \\
    SFT-LoRA & agent & 61.2(-6.1) & 73.0(-1.6) & 84.5 & 80.6(-1.6) \\
    SFT &  mixed & 62.0(-5.3) & 73.0(-1.6) & 84.5 & 80.6(-1.6) \\
    SFT-LoRA & mixed & 62.0(-5.3) & 73.0(-1.6) & 84.5 & 80.6(-1.6) \\
    KTO & agent & 65.3(-2.0) & \textbf{76.6}(+2.0) & \textbf{84.6} & \textbf{82.5}(+0.3) \\
    \hline
    \end{tabular}
    }
    \caption{The impact on the UI localization ability of different finetuning approaches. The experiment is conducted using the UI-Grounding prompt template for all models.}
    \label{fig:ui_acc}
\end{table}

\vspace{1mm}
\noindent \textbf{Analysis of multi-round KTO.} 
We compare the multi-round KTO training with the SFT training on three categories of tasks from the WinAgentArena benchmark: File Explorer, Web Browser, and VsCode, illustrated in Fig.~\ref{fig:ablation_split}. The results show that SFT performs comparably to the OmniParser baseline with GPT-4o, but KTO consistently improves task success rates across rounds. For instance, in the File Explorer split, KTO reaches a $46$\% success rate by the third round (R3), outperforming SFT and OmniParser. Similarly, in the Web Browser and VsCode splits, KTO steadily boosts performance, with the R3 agent achieving $26$\% and $18$\% success rates, respectively. These results highlight the effectiveness of multi-round KTO in enhancing agent performance across different task domains.

\vspace{1mm}
\noindent \textbf{Cost and efficiency analysis.} The cost-efficiency analysis reveals a significant improvement in both time and inference cost when using our 7B grounding model and agent, compared to OmniParser. Our agent model achieves a processing time of 0.4 seconds per frame at a cost of \$6 per 1,000 tasks, vastly outperforming OmniParser, which takes 32 seconds per frame at a cost of \$530. Additionally, our grounding model with the GPT-4o planner not only surpasses OmniParser with a $3.5$\% higher task success rate but also delivers a 10x speed improvement.

\begin{table}[ht]
    \centering
    \resizebox{\linewidth}{!}{
    \begin{tabular}{cccc}
         \hline
         Method & Model Size & Time(s/frame) &  Cost(\$/1,000 tasks) \\
         \hline
         OmniParser & GPT-4o & 32 & $530$\\
         Ours-Ground & GPT-4o & 2.4 & $430$ \\ 
         Ours-Agent & 7B & 0.4 & $6$ \\
         \hline
    \end{tabular}
    }
    \caption{The time and inference cost for different methods. Ours-Ground means our UI-Grounding model with the GPT-4o planner. We use the API pricing of LLama3 8B to measure the cost of our agent model.}
    \label{tab:ui_sft}
\end{table}

\section{Conclusions}


In this work, we presented STEVE, a scalable step verification pipeline aimed at improving the training of computer-use agents. By integrating GPT-4o as a step verifier, STEVE generates a comprehensive trajectory dataset with fine-grained, stepwise reward signals. We further employ KTO to optimize the agent's performance given the binary step verification results. Our experiments showed that KTO effectively leverages both positive and negative examples from the trajectory data, enabling the agent to generalize better and avoid the degradation in UI localization precision observed with SFT. Additionally, we observed that as the number of collected trajectories increased, the performance of our KTO-trained agent consistently improved, underscoring the scalability of our approach. Our results highlight the potential of STEVE to significantly enhance the efficiency and effectiveness of training computer-use agents, particularly in complex real-world desktop environments.


\clearpage
\setcounter{page}{1}
\maketitlesupplementary
\appendix

In this supplementary material, we provide more details about the training settings in Sec.~\ref{sec:implementation}. In Sec.~\ref{sec:prompt}, we present the detailed prompts for our computer-use agents, GPT-4o step verifier, and the GPT-o1 task generator, whereas in Sec.~\ref{sec:demo}, we showcase qualitative results of our agent.

We strongly encourage the readers to explore the videos and the agent trajectories provided in the GitHub repo. These materials offer high-resolution 1080P screenshot inputs, detailed prompts, and complete model responses. 

\setlength{\cftbeforesecskip}{0.5em}
\cftsetindents{section}{0em}{1.8em}
\cftsetindents{subsection}{1em}{2.5em}
\cftsetindents{subsubsection}{3.0em}{3.5em}
\renewcommand{\cftdotsep}{0.5}


\renewcommand{\cftdotsep}{0.5}

\renewcommand{\contentsname}{\large Appendix Contents}

\hypersetup{linkbordercolor=black, linkcolor=black}
\localtableofcontents
\hypersetup{linkbordercolor=link, linkcolor=link}

\tcbset{
  promptstyle/.style={
    colback=gray!5!white, 
    colframe=black,       
    width=\linewidth,     
    boxrule=1pt,          
    arc=4mm,              
    left=5pt,             
    right=5pt,            
    top=5pt,              
    bottom=5pt,           
  }
}

\newenvironment{Prompt}[1]{%
  \begin{tcolorbox}[promptstyle, title={#1}]
}{%
  \end{tcolorbox}
}
\newcommand{\plaincaption}[1]{%
  \vspace{0.5em} 
  \noindent\textbf{#1} 
  \par 
}

\section{Implementation Details}
\label{sec:implementation}
In this section, we delve into the experimental details of the proposed STEVE framework.
We adopt Qwen2-VL~\cite{wang2024qwen2} 7B as the base vision language model for the UI-grounding model. We further fine-tune the agent models from the UI-grounding model.

\subsection{Training Details}
The specifics of our UI-grounding model and KTO agent implementation are given in Tab.~\ref{tab:train_param}.

\begin{table}[ht]
\centering
\resizebox{\linewidth}{!}{
    \begin{tabular}{lclc}
    \toprule
    \multicolumn{2}{c}{\textbf{UI-grounding}} & \multicolumn{2}{c}{\textbf{KTO Agent}} \\
    \cmidrule(r){1-2} \cmidrule(l){3-4}
    \textbf{Config} & \textbf{Value} & \textbf{Config} & \textbf{Value} \\
    \midrule
    base model & Qwen2-VL & base model & UI-grounding \\
    optimizer & AdamW & optimizer & AdamW \\
    scheduler & Cosine & scheduler & Cosine \\
    learning rate & 2e-5 & learning rate & 5e-5 \\
    training data & grounding & training data & agent \\    
    batch size & 32 & batch size & 16 \\
    epochs & 1 & epochs & 2 \\
    vision encoder & freeze & vision encoder & freeze \\
    \bottomrule
    \end{tabular}
    }
    \caption{Settings of our UI-grounding model (left) and KTO agent training (right).}
    \label{tab:train_param}
\end{table}
Specifically, we introduce LoRA~\cite{hu2021lora} during the KTO~\cite{ethayarajh2024kto} training to reduce the memory overhead for the reference and policy model. The settings of the KTO training are outlined in Tab.~\ref{tab:kto_param}. 


\begin{table}[ht]
\centering
    \begin{tabular}{lc}
        \hline
        Config & Value \\
        \hline 
        LoRA $r$ &  256 \\
        LoRA $\alpha$ & 16 \\
        LoRA modules & linear layers in LLM \\
        KTO $\beta$ & 0.1 \\
        KTO $\lambda_D$ & 1.0 \\
        KTO $\lambda_U$ & 1.0 \\
        \hline
    \end{tabular}
    \caption{KTO and LoRA hyperparameters.}
    \label{tab:kto_param}
\end{table}

\subsection{KTO Reward Margin}
\begin{figure}[t]
\centering
    \includegraphics[width=0.9\linewidth]{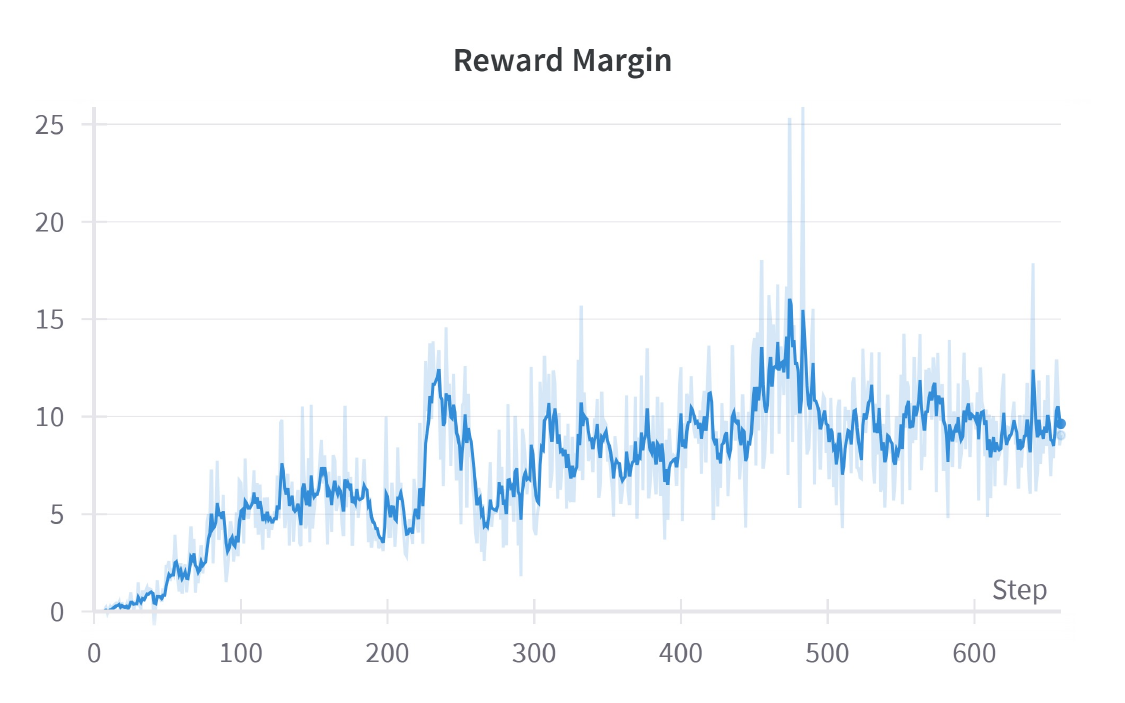}
    \caption{
       The reward margin (vertical axis) between the chosen and rejected samples consistently improve during the KTO training.
    }
\label{fig:reward_margin}
\end{figure}

\vspace{1mm}
The plot in Fig.~\ref{fig:reward_margin} presents the reward margin between the chosen and rejected samples during the KTO optimization. The reward margin steadily increases throughout the training process, indicating that the model performance consistently improves in distinguishing between the desired and undesired actions in the sampled trajectories. 

\section{Prompt Examples}
\label{sec:prompt}
In this section, we will introduce the prompts we used for (I) the computer-use agents, (II) the GPT-4o step verifier, and (III) the GPT-o1 task generator. 

\subsection{Computer-use Agent Prompts}
We provide the prompt for our computer-use agent in Tab.~\ref{tab:agent}.
For our UI-grounding model with the GPT-4o planner, we include more examples in the prompt for the GPT-4o to have a comprehensive understanding of the action space, as proposed by the Navi agent~\cite{bonatti2024windows}. 


\begin{table*}
    \begin{tcolorbox}[title=Computer-use Agent Prompt]
    \footnotesize 
      You are Screen Helper, an AI that executes code to complete tasks on a user's computer. Follow these guidelines: \\
1. Plan efficiently with minimal steps. \\
2. Execute one action per step, unless you're done. \\
3. Verify progress after each step, using the previous image and actions. \\
4. Do not repeat task instructions or screen content in your responses. \\

Input: \\
1. User objective: Task goal (constant until completion) \\
2. Screen images \\
3. History information: Previous N actions/decisions/screen analysis \\
4. Additional human context: User-provided information \\

Available functions: \\
\texttt{```}python
\vspace{-\baselineskip}
\begin{verbatim}
# GUI functions 
computer.mouse.move("element") # Use it frequently 
computer.mouse.single_click() 
computer.mouse.double_click() 
computer.mouse.right_click()  
computer.mouse.scroll(dir="up/down") 
computer.mouse.drag("element") 

# Keyboard functions, similiar to the pyautogui format. 
computer.keyboard.write("text")
computer.keyboard.press("key") 
computer.keyboard.keyDown("key")
computer.keyboard.keyUp("key") 
computer.keyboard.hotkey("key1", "key2", ...) 
\end{verbatim}
\vspace{-\baselineskip}
\texttt{```} \\

Important reminders: \\
1. Pay attention to all fields specified in the task and visible on the screen. \\
2. Extract and address all required fields from the user's intent. \\
3. Verify task completion before sending DONE. \\
4. Avoid repeating unsuccessful actions. \\
5. Execute only one action per step to ensure accuracy and proper interaction with dynamic elements. \\
6. Always analyze the screen content and action history to make correct decisions. \\
7. If you spot a mistake in the last action(mis-click or invalid click), correct it in the current step. \\
8. Write python code to solve complex tasks efficiently. \\
9. For text file and document editing tasks, always select and replace the entire text. Use hotkeys ctrl+a and keyboard.write to avoid partial text selection. \\
10. If you need extra information(for example, personal information to fill a form) to complete the task, set the decision to "FAIL". \\

Input screenshot: \texttt{<image>} \\
User objective: \texttt{<task>} \\
History information: \texttt{<history>} \\
    \normalsize
\end{tcolorbox}
\clearpage

\caption{The prompt for the computer-use agents.}
\label{tab:agent}
\end{table*}

\subsection{GPT-4o Step Verifier Prompts}
We present the prompt for the GPT-4o step verifier in Tab.~\ref{tab:verifier}.
We ask the GPT-4o to observe the screens before and after an action is executed and determine whether the action is beneficial or harmful for the user task completion. 

\begin{table*}
\begin{tcolorbox}[title=GPT-4o Step Verifier Prompt]
\footnotesize 
You are a screenshot data annotator. The user executed some actions on a computer. Your job is to determine whether the actions are beneficial or harmful to the user's objective. \\

You are shown a screenshot, an intermediate action that the user took on the screenshot to complete the goal, and the screen result after the action was taken. You should determine if the action was beneficial or harmful to the user's objective and provide a reason. If you are unsure, you can annotate the action as neutral. The user's action is represented as python functions: \\

\texttt{```}python
\vspace{-\baselineskip}
\begin{verbatim}
# GUI functions 
computer.mouse.move("element") # Use it frequently 
computer.mouse.single_click() 
computer.mouse.double_click() 
computer.mouse.right_click()  
computer.mouse.scroll(dir="up/down") 
computer.mouse.drag("element") 

# Keyboard functions, similiar to the pyautogui format. 
computer.keyboard.write("text")
computer.keyboard.press("key") 
computer.keyboard.keyDown("key")
computer.keyboard.keyUp("key") 
computer.keyboard.hotkey("key1", "key2", ...) 
\end{verbatim}
\vspace{-\baselineskip}
\texttt{```} \\

The annotation guidelines to follow are: \\
1. First analyze whether the screen analysis, multi-step plan, and the current action are reasonable to the user's objective. If there are mistakes in screen analysis, the multi-step plan, or the current action for the task, you should note the sample as harmful. \\
2. If the action is reasonable, you should observe the screen before and after the action to verify if the action is beneficial or harmful and whether the action achieves the expected result. If not, you should mark it as harmful. \\
3. Spot the mistakes in the user's clicking and typing if the action is reasonable. The user might mis-click or type in the wrong position. You should spot the mistakes in the annotation and annotate the action as harmful. The clicking or typing position is drawn in the screen image as a red bounding box. \\
\texttt{```}python
\vspace{-\baselineskip}
\begin{verbatim}
computer.mouse.move("the file 'example.txt'")
computer.mouse.double_click()
\end{verbatim}
\vspace{-\baselineskip}
\texttt{```} \\
Examples: \\
1. Example of reasonable action but wrong clicking position. \\
The user objective: Open the file named "example.txt" under the folder "Documents".
The current screen displays the file explorer with the file "example.txt" highlighted. 
The user executed the following action: I found the file "example.txt" in the current file explorer and the next action is double-clicking on it. \\
The red bounding box is drawn around the Quick access folder "Pictures" instead of the file "example.txt". And the result screen shows the content of the folder "Pictures". Then the action is harmful because the user clicked on the wrong position. \\
2. Example of unreasonable action. \\
The user objective: Search for the price of the latest iPhone on eBay. \\
The current screen shows the search results of the "latest iPhone" on eBay. \\
The user executed the following action: To find the price of the latest iPhone, I will type "latest iPhone" in the search bar. \\
The action is harmful because the user is already on the search results page of "latest iPhone" on eBay. The correct action should be clicking on the product link to view the details. \\
\\
More examples. \\
\\
Your analysis of the action taken, the change of the screen, the user objective's status, and the annotation of the action. Then output a JSON object with the following fields: \\
\texttt{```}json
\vspace{-\baselineskip}
\begin{verbatim}
{
    "annotation": "GOOD/NEUTRAL/HARMFUL"
}
\end{verbatim}
\vspace{-\baselineskip}
\texttt{```}
    \normalsize
\end{tcolorbox}
\caption{The detailed prompts for the GPT-4o step verifier.}
\label{tab:verifier}
\end{table*}

\subsection{Task Generation Prompts}
We present the prompts designed for the GPT-o1 model to generate real-world, feasible tasks, as outlined in Tab.~\ref{tab:task}, particularly for the Windows File Explorer tasks.
Following the task configuration format defined in WinAgentArena~\cite{bonatti2024windows}, we prompt GPT-o1 to produce similar tasks. Functions such as creating folders, downloading files, or opening applications are pre-executed to establish a feasible initial state for the agent to complete the assigned task. To support task generation, we compile a collection of document files, image files, and website URLs, which are provided within the prompt for GPT-o1 to utilize in creating practical and executable tasks.

\begin{table*}
\begin{tcolorbox}[title=Task Generation Prompt for the GPT-o1]
\footnotesize 

I want to collect a list of tasks that can be done on a personal computer about basic Windows system operations. \\

Here are some examples of the tasks: \\
1. Rename the file 'file1.txt' to 'file2.txt'.\\
2. Navigate to the folder 'folder1'.\\
3. Move the file 'C://file1.txt' to 'C://target\_folder/file1.txt'. \\
4. Create a shortcut of "C://Users//Administrator//Documents//file1.txt" on the desktop.\\
5. Open the file 'C://Users//Administrator//Downloads//file1.txt'.\\
6. Navigate to the folder 'folder2'. List the names of all the files in the folder and save them as a file 'names.txt' under that folder.\\
7. Close the current window.\\
8. Copy all the files in the folder 'folder1' to the folder 'folder2'.\\
9. Summarize the content of the pdf file 'C://paper.pdf' and save it as 'summary.txt' under the Documents folder.\\

You should avoid creating infeasible tasks. For example, if you want to create a task "rename the file 'C://Users//Administrator//Documents//file1.txt' to 'C://Users//Administrator//Documents//file2.txt'", you should ensure that the file 'file1.txt' exists before creating the task.\\

You have the following tools to make a task feasible:\\
1. create\_folder: create a folder at a specific path.\\
2. download: download files and save them to the local computer.\\
3. sleep: sleep for a specific time.\\
4. open: open a file or a folder.\\
5. launch: launch a folder\\
Here are some examples of the tools:\\
1. Use the 'create\_folder' tool to create a folder named 'Projects' in the Documents folder.
\vspace{-\baselineskip}
\begin{verbatim}
{"type": "create_folder", "parameters": {"path": "C:\\Users\\Administrator\\Documents\\Projects"}}    
\end{verbatim}
\vspace{-\baselineskip}
2. Use the 'download' tool to download the file 'file1.txt' from the internet and save it to the local computer.
\vspace{-\baselineskip}
\begin{verbatim}
{"type": "download", "parameters": {"files": [{"url": "https://file1.txt", 
 "path": "C:\\Users\\Administrator\\Downloads\\file1.txt"}]}}
\end{verbatim}
\vspace{-\baselineskip}
3. Use the 'sleep' tool to sleep for 2 seconds.
\vspace{-\baselineskip}
\begin{verbatim}
{"type": "sleep", "parameters": {"seconds": 2}}
\end{verbatim}
\vspace{-\baselineskip}
4. Use the 'open' tool to open the file 'file1.txt'.
\vspace{-\baselineskip}
\begin{verbatim}
{"type": "open", "parameters": {"path": "C:\\Users\\Administrator\\Downloads\\file1.txt"}}
\end{verbatim}
\vspace{-\baselineskip}
5. Use the 'launch' tool to launch the folder 'Projects'.
\vspace{-\baselineskip}
\begin{verbatim}
{"type": "launch", "parameters": {"path": "C:\\Users\\Administrator\\Documents\\Projects"}}
\end{verbatim}

Here are some examples to create new feasible tasks by setting up the configurations: \texttt{<seed\_tasks>} \\

You have access to the following file URLs and you can download them to a specific path on the local computer and then create tasks:
\vspace{-\baselineskip}
\begin{verbatim}
{
    "cat_image.jpg": "https://cat_image.jpg", # URL for downloading the file
    "example_ppt.pptx": "https://example_ppt.pptx",
    "example_word.docx": "https://example_word.docx",
}
\end{verbatim}
\vspace{-\baselineskip}

Keep the rules in mind:\\
1. You can only use the provided files to generate the tasks.\\
2. Ensure that the tasks are feasible and can be completed on a personal computer by setting up the necessary configurations.\\
3. In order to make the task feasible. You can create folders and files in advance under the Downloads folder('C://Users//Administrator//Downloads') or Documents and Pictures. Make subfolders and files as needed. The download process is defined in the "config" part of the JSON.\\
4. Make sure the URL is correct if you use the download tool.\\
5. Do not mention the preprocessing steps like downloading files in the instructions. Only mention the final task that the user needs to do.\\

Here is an example task, now you are required to generate 10 similar tasks based on this seed task. For example, if the original instruction is "book a flight from Shanghai to Beijing", you can generate a similar one like "What's the price of a flight from LA to New York.". Do not deviate too much from the original instruction. \\
The task to modify: \texttt{<seed task>}. You should output the generated tasks in the following JSON format: \\
\texttt{```}json
\vspace{-\baselineskip}
\begin{verbatim}
{
    "tasks": [
        {"instruction": "xx", "config": {...}},
        {"instruction": "xx", "config": {...}}
    ]
}
\end{verbatim}
\vspace{-\baselineskip}
\texttt{```}
    \normalsize
\end{tcolorbox}
\caption{The task generation prompt for the GPT-o1 model.}
\label{tab:task}
\end{table*}


\section{Agent Demo}
\label{sec:demo}
This section presents visualizations of various agents performing tasks within the WinAgentArena environment. Specifically, it highlights the successful task trajectories of our STEVE-KTO-7B agent in Fig.~\ref{fig:supp-fig1},~\ref{fig:supp-fig2},~\ref{fig:supp-fig3}.
Additionally, the performance of the SFT agent, the KTO agent, and the UI-grounding model is compared with that of GPT-4o.

\subsection{WinAgentArena Examples}
\begin{figure*}[p]
    \centering
    \includegraphics[width=0.95\linewidth]{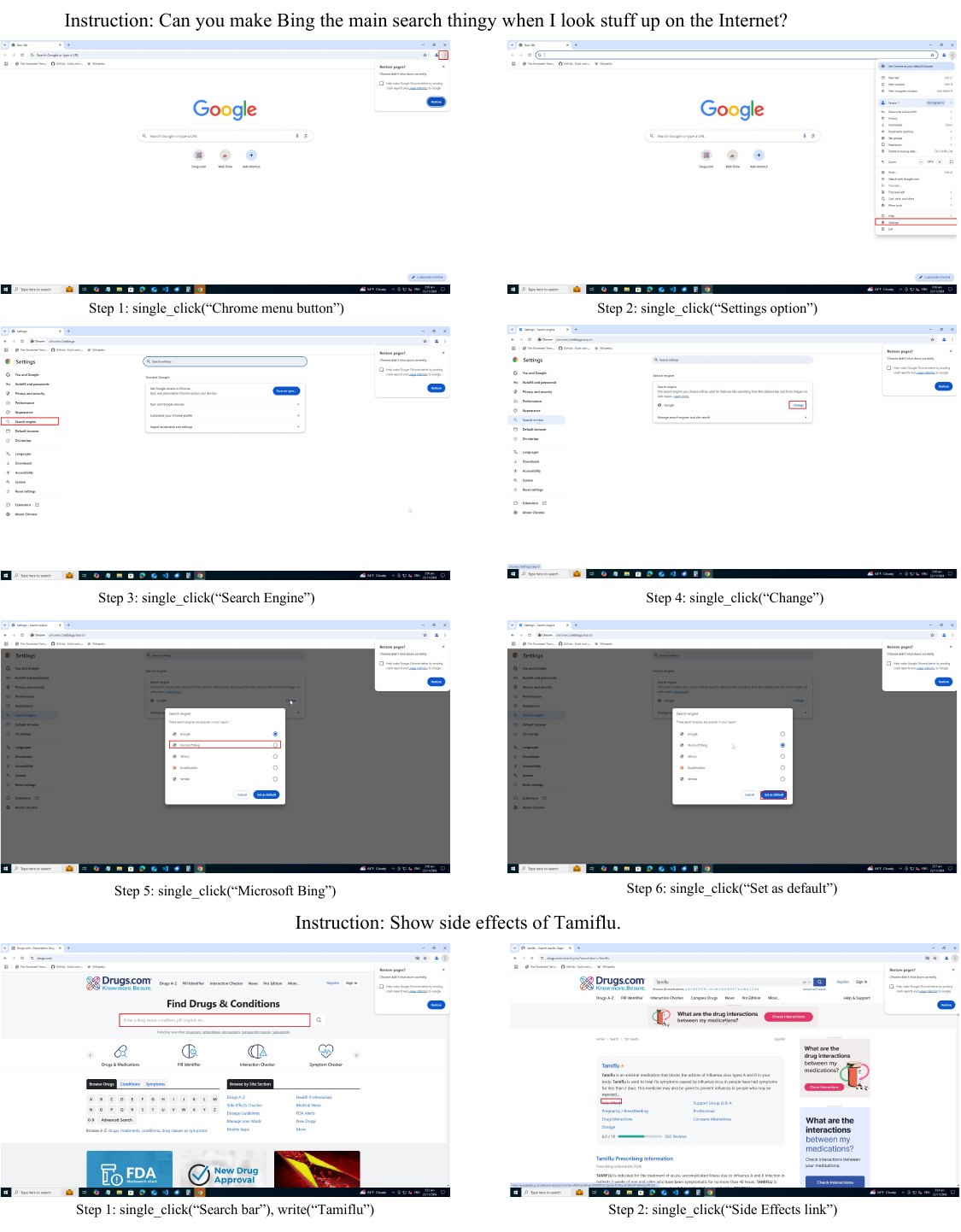}
    \caption{The trajectories of our STEVE-KTO-7B agent for the chrome tasks from the WinAgentArena~\cite{bonatti2024windows} with ID bb5e4c0d-f964-439c-97b6-bdb9747de3f4-wos (up) and b070486d-e161-459b-aa2b-ef442d973b92-wos (bottom). We display a simplified action for each step and plot the target UI localization results with a red bounding box in each screenshot. For high-resolution screenshots/videos, full model responses with screen analysis, multi-step planning, and python code blocks, please refer to the corresponding attachments.}
    \label{fig:supp-fig1}
\end{figure*}

\begin{figure*}[p]
    \centering
    \includegraphics[width=0.92\linewidth]{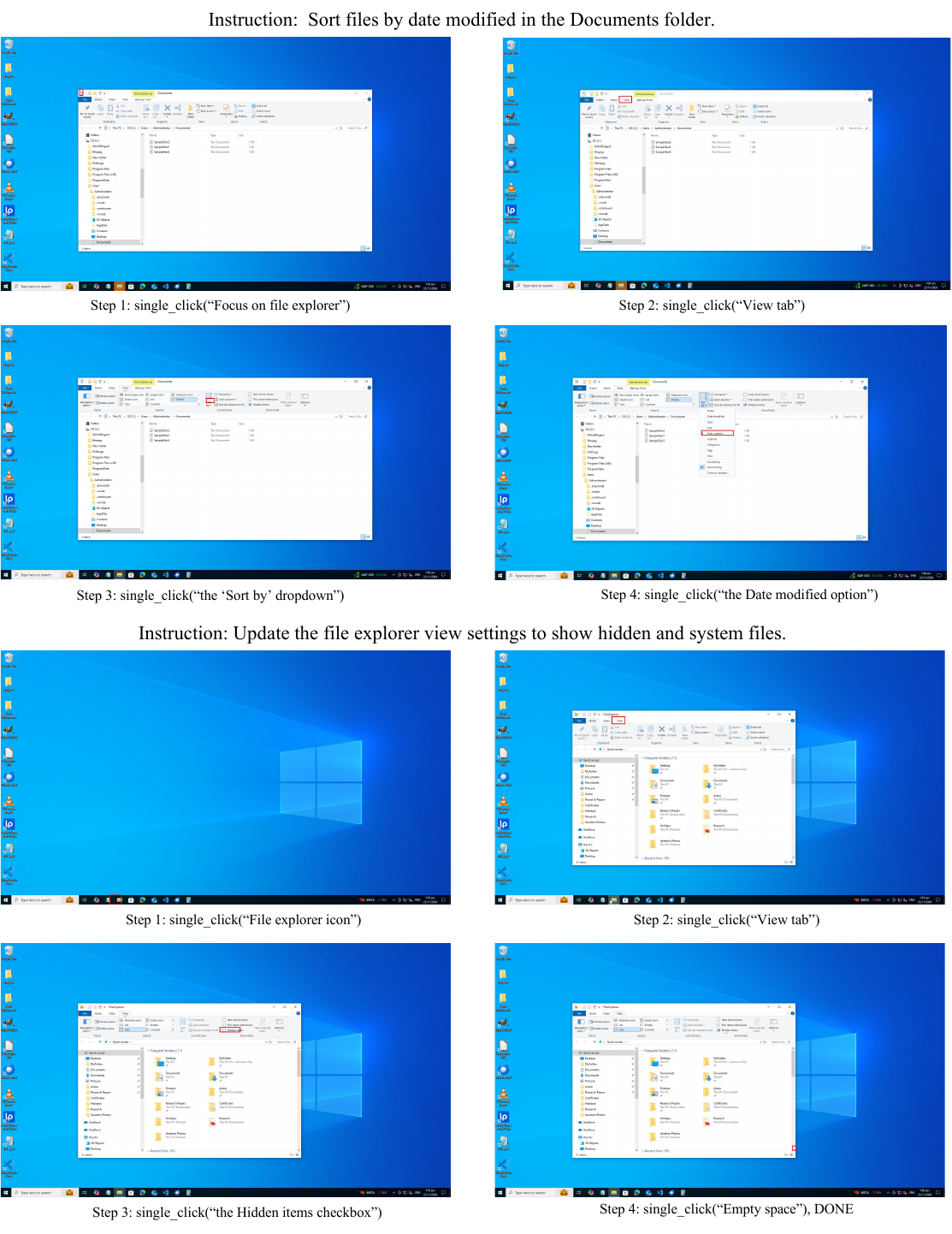}
    \caption{The trajectories of our STEVE-KTO-7B agent for the file explorer tasks from the WinAgentArena~\cite{bonatti2024windows} with ID 7c70e16b-e14f-4baa-b046-3e022b2d0305-WOS (up) and 5316686e-5688-4115-be24-052037df599f-WOS (bottom). We display a simplified action for each step and plot the target UI localization results with a red bounding box in each screenshot. For high-resolution screenshots/videos, full model responses with screen analysis, multi-step planning, and python code blocks, please refer to the corresponding attachments.}
    \label{fig:supp-fig2}
\end{figure*}

\begin{figure*}[p]
    \centering
    \includegraphics[width=0.95\linewidth]{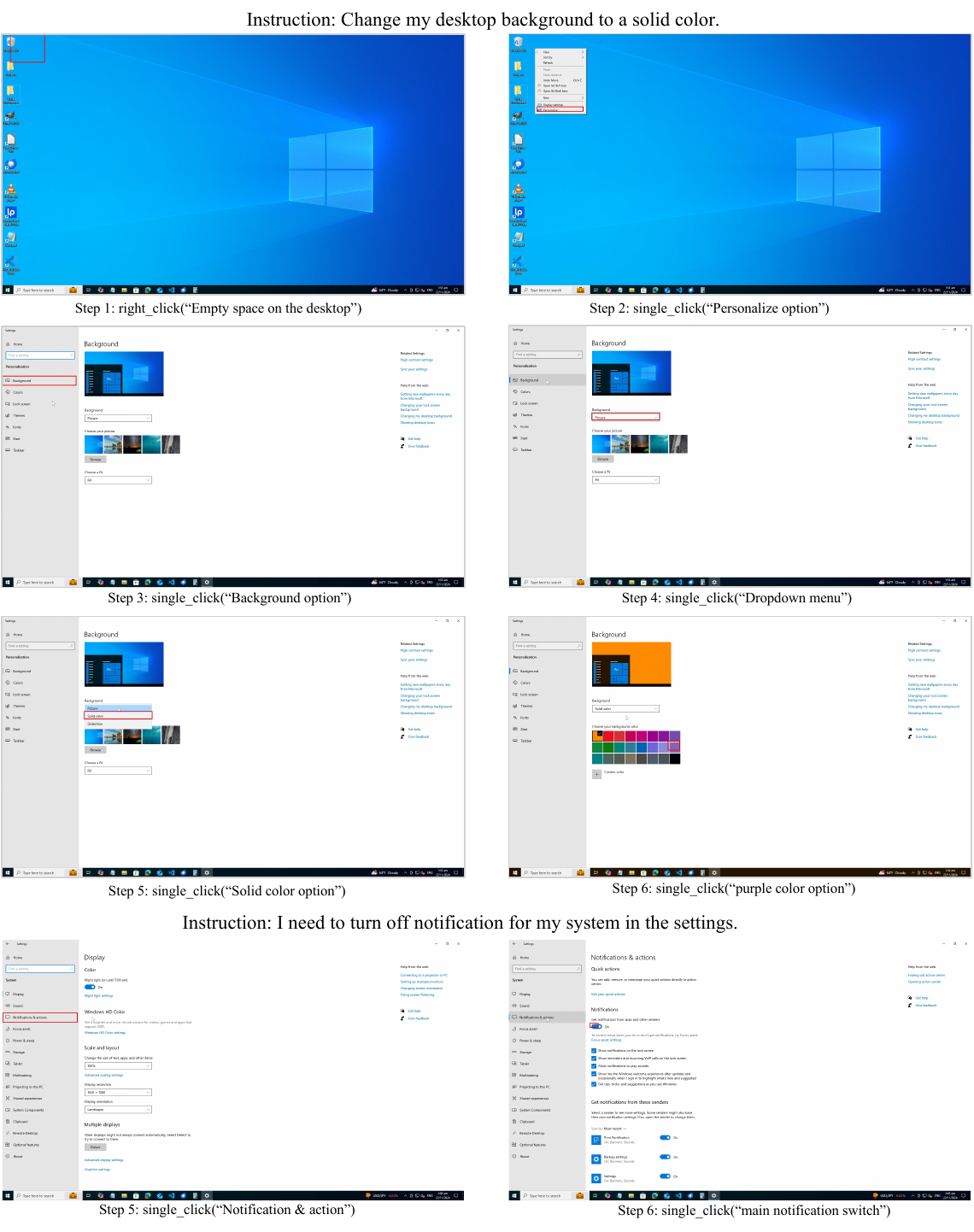}
    \caption{The trajectories of our STEVE-KTO-7B agent for the Windows setting tasks from the WinAgentArena~\cite{bonatti2024windows} with ID a659b26e-4e31-40c1-adaf-34742b6c44ac-wos (up) and 37e10fc4-b4c5-4b02-a65c-bfae8bc51d3f-wos (bottom). Only the last two steps of the later task are shown. We display a simplified action for each step and plot the target UI localization results with a red bounding box in each screenshot. For high-resolution screenshots/videos, full model responses with screen analysis, multi-step planning, and python code blocks, please refer to the corresponding attachments.}
    \label{fig:supp-fig3}
\end{figure*}

In Fig.~\ref{fig:supp-fig1},~\ref{fig:supp-fig2},~\ref{fig:supp-fig3}, we present the successful tasks of our STEVE-KTO 7B agent on the Chrome browser, file explorer, and Windows setting tasks from the WinAgentArena benchmark. For a more comprehensive visualization, we encourage readers to view the screen recordings or examine the agent trajectories provided in the HTML logs.

\subsection{Comparisons between agents}
\begin{figure*}[p]
    \centering
    \includegraphics[width=0.72\linewidth]{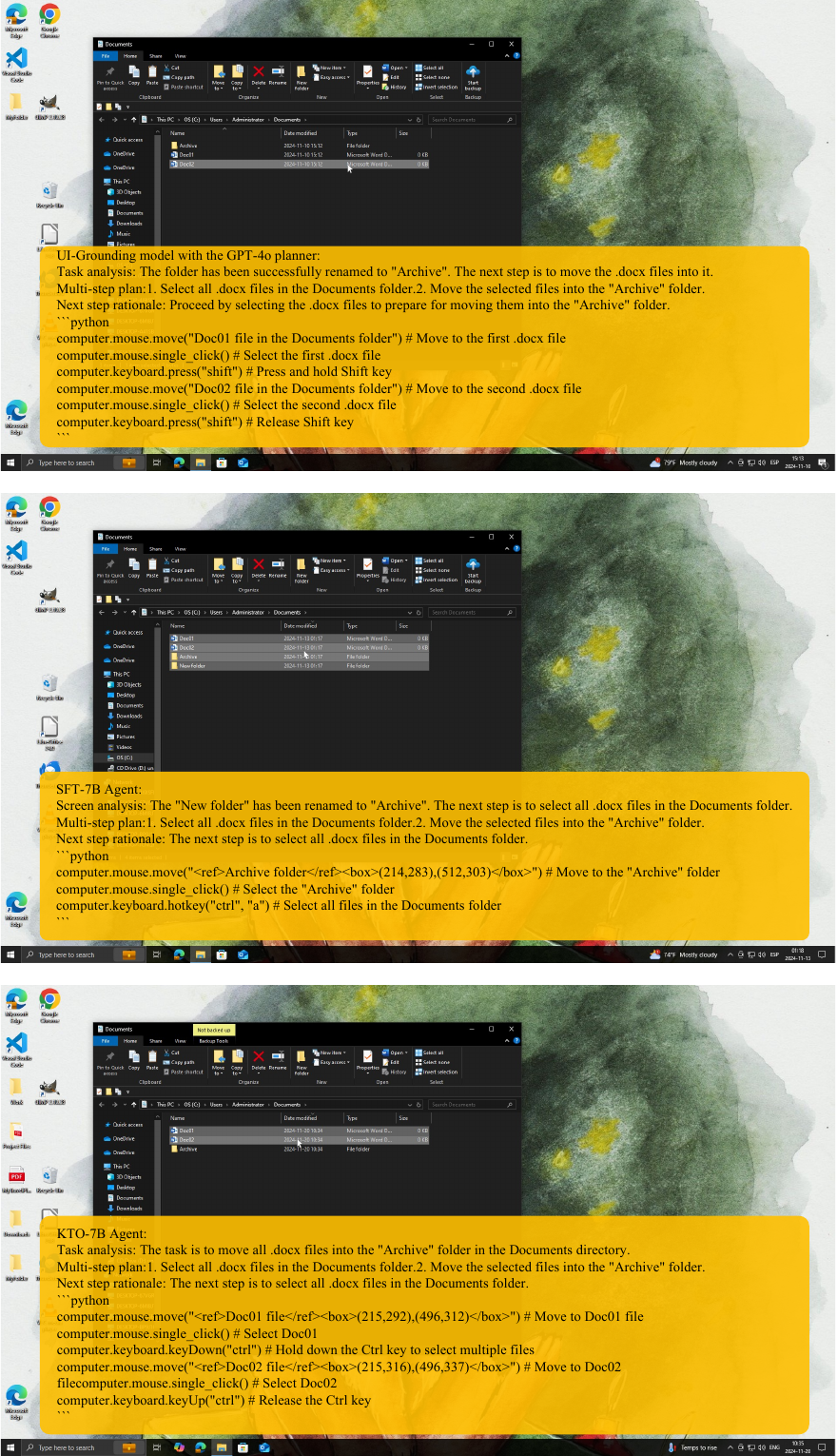}
    \caption{Comparisons of different computer-use agent models.}
    \label{fig:supp-compare}
\end{figure*}

Although the UI-grounding model with the GPT-4o as its planner achieves the best overall performance on the WinAgentArena benchmark, we found that the KTO 7B agent outperforms GPT-4o in certain tasks. Fig~\ref{fig:supp-compare} presents the behaviors of different agents for the same instruction, \textit{``Move the document files into the Archive folder''}. The GPT-4o planner made a correct high-level decision to select all the docx files. However, due to a lack of comprehensive understanding of the action space, it utilized ``press'' instead of ``keyDown'' for the task, leading to a file missed in the selection. The SFT-agent tried to select all files by the hotkey ``Ctrl+A'', which included extra contents for the cut and paste operation. In contrast, the KTO-agent successfully select two docx files by a sequence of correct mouse and keyboard actions. We attached more successful task trajectories of the KTO-agent model in the materials that demonstrate the effectiveness of our step verification training pipeline.  

{
    \small
    \bibliographystyle{ieeenat_fullname}
    \bibliography{main}
}

\end{document}